\pdfoutput=1

\documentclass[11pt]{article}

\usepackage{emnlp2021}

\usepackage{times}
\usepackage{latexsym}

\usepackage{hyperref}
\usepackage{multirow}
\usepackage{booktabs}
\usepackage[T1]{fontenc}

\usepackage[utf8]{inputenc}

\usepackage{microtype}

\usepackage{comment}

\usepackage{hhline}

\usepackage{graphicx}
\usepackage{caption}
\usepackage{subcaption}

\usepackage{amsmath, amsbsy, bm}
\newcommand*\bell{\ensuremath{\boldsymbol\ell}}
\usepackage{amssymb}

\usepackage{threeparttable}

\usepackage{todonotes}

\usepackage{colortbl}

\usepackage{xr}
\usepackage{cleveref}

\usepackage{amssymb}
\usepackage{pifont}
\newcommand{\cmark}{\ding{51}}%
\newcommand{\xmark}{\ding{55}}%

\makeatletter
\newcommand*{\addFileDependency}[1]{
  \typeout{(#1)}
  \@addtofilelist{#1}
  \IfFileExists{#1}{}{\typeout{No file #1.}}
}
\makeatother

\title{Skim-Attention: Learning to Focus via Document Layout}

\author{Laura Nguyen $^{\diamond\star}$ \quad Thomas Scialom$^{\diamond\star}$ \quad Jacopo Staiano$^{\star}$ \quad Benjamin Piwowarski$^{\diamond}$\\
$^\diamond$ Sorbonne Universit\'e, CNRS, LIP6, F-75005 Paris, France\\
$^\star$ reciTAL, Paris, France \\
  {\tt \{laura,thomas,jacopo\}@recital.ai} \\
  {\tt benjamin.piwowarski@lip6.fr}\\}

\begin{document}
\maketitle

\begin{abstract}

Transformer-based pre-training techniques of text and layout have proven effective in a number of document understanding tasks. Despite this success, multimodal pre-training models suffer from very high computational and memory costs. Motivated by human reading strategies, this paper presents \emph{Skim-Attention}, a new attention mechanism that takes advantage of the structure of the document and its layout. \emph{Skim-Attention} only attends to the 2-dimensional position of the words in a document. Our experiments show that \emph{Skim-Attention} obtains a lower perplexity than prior works, while being more computationally efficient. \emph{Skim-Attention} can be further combined with long-range Transformers to efficiently process long documents. We also show how  \emph{Skim-Attention} can be used off-the-shelf as a mask for any Pre-trained Language Model, allowing to improve their performance while restricting attention. Finally, we show the emergence of a document structure representation in \emph{Skim-Attention}.

\end{abstract}

\section{Introduction}

More and more companies have started automating their document processing workflows by leveraging artificial intelligence techniques. This had lead to the emergence of a dedicated research topic, Document Intelligence\footnote{\url{https://sites.google.com/view/di2019}} (DI), which encompasses the techniques used to read, interpret and extract information from business documents. Such documents span multiple pages and contain rich multi-modal information that include both text and layout. Earliest approaches to analyzing business documents rely on rule-based algorithms \citep{lebourgeois1992fast, amin2001page}, but the success of deep learning has put computer vision and natural language processing (NLP) models at the heart of contemporary approaches \citep{katti2018chargrid, denk2019bertgrid}. With the massive impact of large pre-trained Transformer-based language models \citep{devlin-etal-2019-bert, radford2019language}, DI researchers have recently started leveraging Transformers.

At the core of the Transformer architecture is self-attention, a powerful mechanism which contextualizes tokens with respect to the whole sequence. While being the key to the success of Transformers, it is also its bottleneck: the time and memory requirements of self-attention grow quadratically with sequence length. As a consequence, only short sequences can be processed (512 tokens or 1,024 at most), making it impossible to capture long-term dependencies. This is an important issue for DI since texts can be very dense and long in business documents. To allow efficient training on very long sequences, there has been growing interest in building model architectures that reduce the memory footprint and computational requirements of Transformers \citep{dai-etal-2019-transformer, kitaev2020reformer, beltagy2020longformer}. This plethora of long-range Transformers lie in one specific research direction: capturing long-range dependencies by reducing the cost of self-attention.

These Transformer architectures all operate on serialized texts, i.e. one-dimensional sequences of words, completely disregarding the document layout. However, layout, i.e. the physical organization of a document's contents, carries useful information about the semantics of the text and has a significant impact on readers’ understanding \citep{wright1999psychology}. Thus, ignoring the document layout leads to a considerable loss of information. To address this issue, an orthogonal direction that has gained traction recently is based on integrating layout information into Transformer-based language models. Joint pre-training of text and layout has allowed models to reach state-of-the-art performance in several downstream tasks concerning layout-rich documents \cite{xu2020layoutlm, pramanik2020towards, xu2020layoutlmv2}. Despite their effectiveness, these approaches all "read" the contents token by token to compute attention. We claim that one does not need to have read each word in a document page to be able to understand a specific paragraph. Thus, we argue that, to efficiently process long documents, it is a waste of effort and computation to contextualize a token with respect to the entire input sequence.

To shift towards processing long documents with awareness of their structure, we propose to take into account layout in a more intuitive and efficient way. First, we present a quick cognitive experiment wherein we show that layout plays a fundamental role in humans’ comprehension of documents. In light of this experiment, we claim that one can already gather a lot of information from the layout alone. As a consequence, we propose Skim-Attention, a new self-attention mechanism that is solely based on the 2-D position of tokens in the page, independently from their semantics. To exploit this mechanism, we introduce Skimformer and SkimmingMask, two frameworks for integrating Skim-Attention into Transformer models. Skimformer is an end-to-end Transformer language model that replaces self-attention with Skim-Attention. Based on the tokens’ spatial locations, Skimformer computes the Skim-Attention scores only once, before using them in each layer of a text-based Transformer encoder. Skimformer can also be adapted to long-range Transformers to model longer documents. Conversely, SkimmingMask uses Skim-Attention as a mask to sparsify attention in any Transformer language model. Each token is restricted to its $k$ most attended tokens, as indicated by Skim-Attention, which allows for a smaller context length.

In summary, our main contributions are as follows:
\begin{itemize}
    \itemsep-8pt
    \item We introduce Skim-Attention, a new attention mechanism that leverages layout.
    \item We design two frameworks for integrating Skim-Attention into Transformer models, and show that they are more time and memory efficient than LayoutLM.
    \item To the best of our best knowledge, this is the first time layout is considered as a means for reducing the cost of self-attention. 
\end{itemize}

\section{Related Work}

\subsection{Cognitive Background}

The layout of a document, which refers to the arrangement and organization of its visual and textual elements, has a significant influence on readers’ behavior and understanding \citep{wright1999psychology, kendeou2007effects, olive2017processing}. It has been shown that a well-designed layout results in less cognitive effort \citep{britton1982effects, olive2017processing} and facilitates comprehension of the conveyed information by helping identify the document type and its constituents, as well as providing cues regarding relationships between elements \citep{wright1999psychology}. Semiotic research assumes that readers scan the document before taking a closer look at certain units \citep{kress1996reading}, a claim supported by eye-tracking experiments on newspapers \citep{leckner2012presentation}. For all these reasons, layout is a critical element for document understanding, which motivates its integration into modeling. Inspired by these research findings, our work focuses on exploiting layout in a similar fashion as humans, since this can be key to a successful model coping with long and complex documents.

\subsection{Long-range Transformers}

In the field of natural language processing, Transformers have become the go-to component in the modern deep learning stack. In recent years, there has been a substantial growth in the number of Transformer variants (\textit{long-range Transformers}) that improve computational and memory efficiency, making it possible to extend the maximum sequence length and to incorporate long-term context. Models such as Longformer \citep{beltagy2020longformer}, Reformer \citep{kitaev2020reformer}, and Performer \citep{choromanski2020rethinking} are able to process sequences of thousands of tokens or longer. Although these models are highly efficient in reducing time and memory requirements, they consider long documents as huge one-dimensional blocks of texts: Reformer, for instance, has to read the 4,096 elements contained in the input sequence in order to create buckets of similar elements. Hence, all information about the document structure is lost.

Our approach is orthogonal to long-range Transformers; instead of focusing only on architecture optimization, we propose to leverage layout-rich information. 

\subsection{Multi-modal Pre-training Techniques for Document Understanding}

Recently, multi-modal pre-training techniques have become increasingly popular in the document understanding area \citep{xu2020layoutlm, pramanik2020towards, garncarek2020lambert, wu2021lampret}. This research direction consists in jointly pre-training on textual and layout/visual information from a large and heterogeneous collection of unlabeled documents, learning cross-modal interactions in an end-to-end fashion. Based on the BERT architecture, \citet{xu2020layoutlm} build LayoutLM, a multi-modal Transformer model that ties spatial information with tokens through a point-wise summation to learn pre-trained embeddings for document understanding tasks.

LayoutLM, along with most approaches in prior-art, is not motivated by efficiency and cognitive perspectives. The layout information is rather considered as an additional feature, and this approach requires to "read" each individual token one by one. As opposed to LayoutLM, in our proposed approach, attention is computed exclusively on spatial positions. This leads to improvements on time and memory efficiency. In addition, our approach can be plugged into any textual language model, making it more flexible than LayoutLM, which requires both text and layout to be learnt jointly in an extensive pre-training stage.

\section{Preliminary Experiments: Human Evaluation}
\label{section:human-eval}

How much does the document layout help in comprehending long textual contents? How faster is it for humans to find information in documents when layout is provided? To answer these questions, we conduct a simple cognitive experiment wherein we measure the amount of time needed for human annotators to retrieve information from both formatted and plain-text documents. Half of the time, they are given access to the full layout, and the other half, to plain text only (i.e., no layout nor formatting).\footnote{For additional details regarding the experimental protocol, documents, questions and results, see Section~\ref{supp-section:human-eval} in the appendix.}

Table~\ref{tab:human-eval} reports the average time needed to retrieve information from the documents. We find that it is $2.5 \times$ faster to answer questions from the formatted documents, and that the variability in the results is much lower in this case. These results support the hypothesis that \emph{less cognitive effort} is spent when the document is formatted, emphasizing the importance of layout information in reading comprehension.

We believe that machines could benefit from the the document layout, just like humans, as a strategy to retrieve information faster while expending less effort. In particular, layout information could be of great help in reducing the cost of self-attention in Transformer models.

\begin{table}
\centering \small
\begin{tabular}{crr}
    \hline
               & \textbf{Average} & \textbf{Standard Deviation} \\
    \hline 
    Formatted  & 6.05  & 1.73 \\
    Plain-text & 15.18 & 9.06 \\
    \hline
\end{tabular}
\caption{Average (std) time (in seconds) required to answer questions from documents, depending on whether layout is provided.}
\label{tab:human-eval}

\end{table}

\section{Proposed Approach}

Using common sense, and in light of the cognitive experiment previously reported, it is clear that thelayout is of utmost importance for humans to understand long documents. We propose to take into account layout by introducing \textit{Skim-Attention}, a self-attention module that computes attention solely based on spatial positions. To process long and layout-rich documents, we propose different ways of integrating this mechanism into Transformer architectures.

\subsection{Background on Transformers}

We first provide an overview of the well-established Transformer, an encoder-decoder architecture composed by stacking a series of Transformer blocks on top of each other. 
Each block is characterized by a self-attention module. Given an input sequence encoded as a matrix $\textbf{X} \in \mathbb{R}^{n \times d}$, the operation for a single layer is defined as:

\begin{equation}
    \bm{\alpha} = \text{Softmax}\left(\dfrac{\bf{Q}\bf{K}^\top}{\sqrt{d}}\right)\bf{V}
\end{equation}

\noindent where $\bf{Q}$, $\bf{K}$ and $\bf{V}$ are the Query, Key and Value matrices obtained by a linear transformation of $\bf{X}$. More intuitively, the attention matrix, $\bf{A} = \bf{Q}\bf{K}^\top$, provides text-based similarity scores for all pairs of tokens in the sequence, while each row in $\text{Softmax}\left(\dfrac{\bf{Q}\bf{K}^\top}{\sqrt{d}}\right)$ represents a distribution that indicates how we need to aggregate information from the input tokens ($\bf{V}$) for the corresponding output token ($\bf{Q}$).

It is clear that the main limitation of Transformers lies in the computational and memory requirements of the attention: to obtain the attention matrix, inner products between each key and each query need to be computed, resulting in a quadratic complexity w.r.t. the input sequence length.
This operation is repeated at each layer, hence processing longer sequence quickly becomes computationally challenging.
Finally, the standard Transformer architecture considers documents as serialized sequences of texts, leading to a severe loss of information when it comes to layout-rich documents.

\subsection{Skim-Attention Overview}

Our novel attention mechanism, Skim-Attention, views documents as collections of boxes distributed over a two-dimensional space, i.e., the page. In the following, we provide details on how to encode spatial positions into layout embeddings, before describing our attention module.

\paragraph{Layout Embeddings}

Layout embeddings carry information about the spatial position of the tokens. Following LayoutLM \citep{xu2020layoutlm}, the spatial position of a token is represented by its bounding box in the document page image, $(x_0, y_0, x_1, y_1)$, where $(x_0, y_0)$ and $(x_1, y_1)$ respectively denote the coordinates of the top left and bottom right corners. We discretize and normalize them to integers in $[0, ..., 1000]$. Four embedding tables are used to encode spatial positions: two for the coordinate axes ($x$ and $y$), and the other two for the bounding box size (width and height). The final layout embedding of a token, $\bell \in \mathbb{R}^{d_{\ell}}$, located at position $(x_0, y_0, x_1, y_1)$ is defined by:
\vspace{-0.5cm}

\begin{equation}
\begin{split}
    \bell & = \text{LayoutEmb}_x(x_0) + \text{LayoutEmb}_y(y_0) \\
    & + \text{LayoutEmb}_x(x_1) + \text{LayoutEmb}_y(y_1) \\
    & + \text{LayoutEmb}_w(x_1 - x_0) \\
    & + \text{LayoutEmb}_h(y_1 - y_0) \\
\end{split}
\end{equation}

\paragraph{Skim-Attention}

We propose Skim-Attention, an attention mechanism that leverages document layout in a novel way. As opposed to standard self-attention, Skim-Attention does not depend on the text semantics (i.e. token representations), as it calculates the attention  using \emph{only} the spatial positions of the tokens, i.e. their layout embeddings $\bell$.

Formally, let $\textbf{X}^{\ell} =  \{\bell_0, \bell_1, …, \bell_n\}$ be an input sequence of layout embeddings, and $\textbf{Q}^{\ell} = \textbf{W}^{\ell}_q \textbf{X}^{\ell}, \textbf{K}^{\ell} =  \textbf{W}^{\ell}_k \textbf{X}^{\ell}$, the Queries and Keys obtained by linear transformations of the layout embeddings. For a single attention head, the Skim-Attention matrix is defined by:

\begin{equation}
\label{eq:skim-attention-matrix}
    \bf{A}^{\ell} = \text{Softmax}\left(\dfrac{\bf{Q}^{\ell}\left(\bf{K}^{\ell}\right)^\top}{\sqrt{d^{\ell}}}\right)
\end{equation}

Intuitively, $\bf{A}^{\ell}$ captures the correlation between two tokens based on their spatial positions: the more similar two tokens are in terms of layout embeddings, the higher their attention score. 

Since attention is calculated only once, we want the layout embeddings to be as meaningful as possible. Therefore, to obtain better layout representations, we contextualize them by adding a small Transformer prior to computing Skim-Attention. 

It is possible to combine Skim-Attention with any long-range Transformer, as these approaches are orthogonal. We adapt our approach by computing the corresponding long-range attention only once, based on layout instead of text semantics.

\subsection{Skim-Attention in Transformers}

We investigate two approaches to exploit Skim-Attention:
\emph{i)} \textit{Skimformer}, wherein self-attention is replaced by Skim-Attention; and \emph{ii)} \textit{SkimmingMask}, where an attention mask is built from Skim-Attention and fed to a Transformer language model.

\begin{figure*}[ht]
  \begin{subfigure}[b]{0.51\textwidth}
    \includegraphics[width=\textwidth]{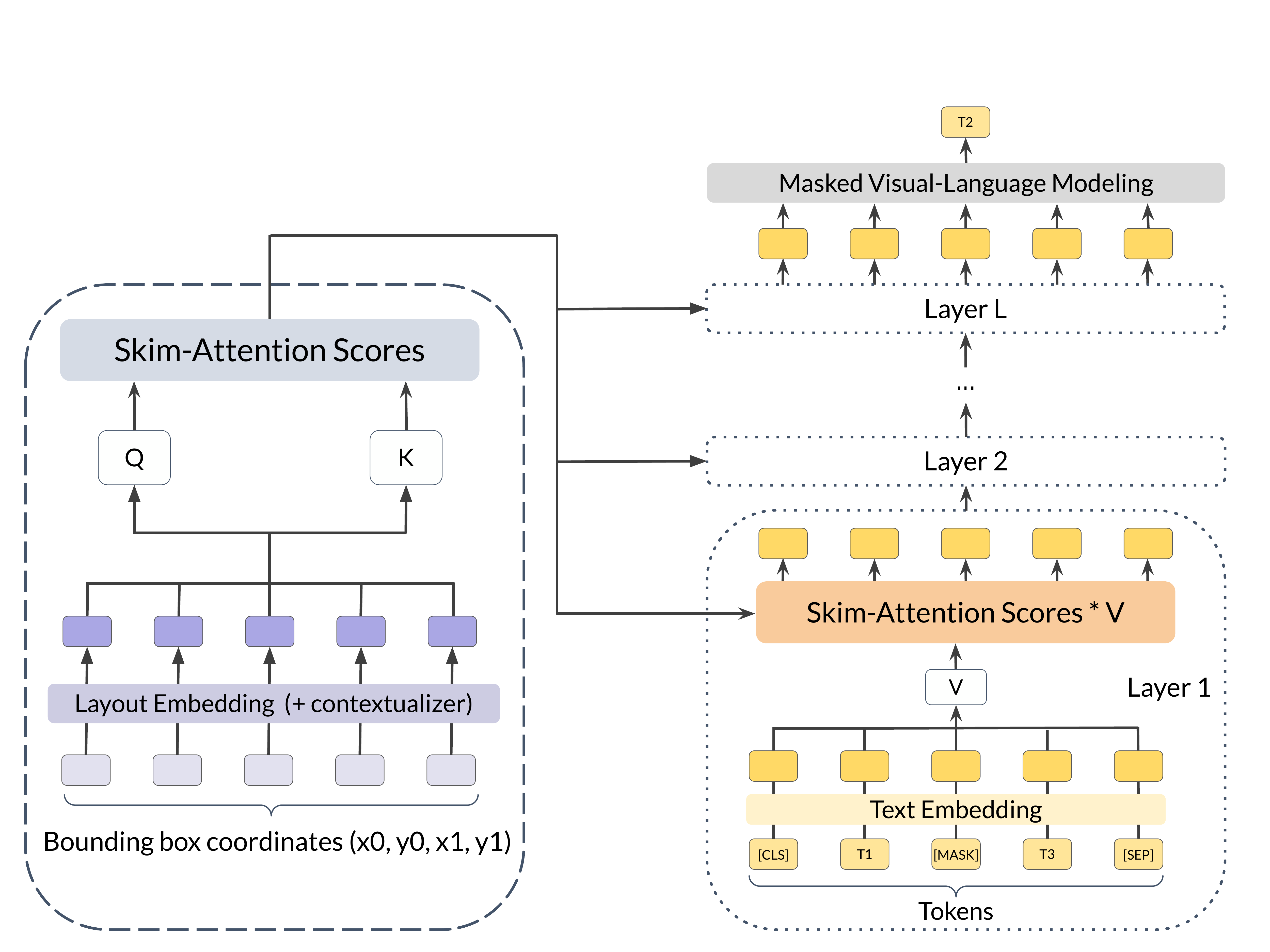}
    \caption{Skimformer model architecture. $L$ denotes the number of Transformer encoder layers. $\textbf{Q}$ and $\textbf{K}$ are the queries and keys obtained by projecting the layout embeddings. $\textbf{V}$ represents the values produced by projecting the encoder layers' textual inputs. The attention is solely based on token spatial positions and computed only once. The attention scores are then distributed to each layer of a Transformer encoder.}
    \label{fig:skimformer-architecture}
  \end{subfigure}
  \hspace{0.25em}
  \begin{subfigure}[b]{0.51\textwidth}
    \includegraphics[width=\textwidth]{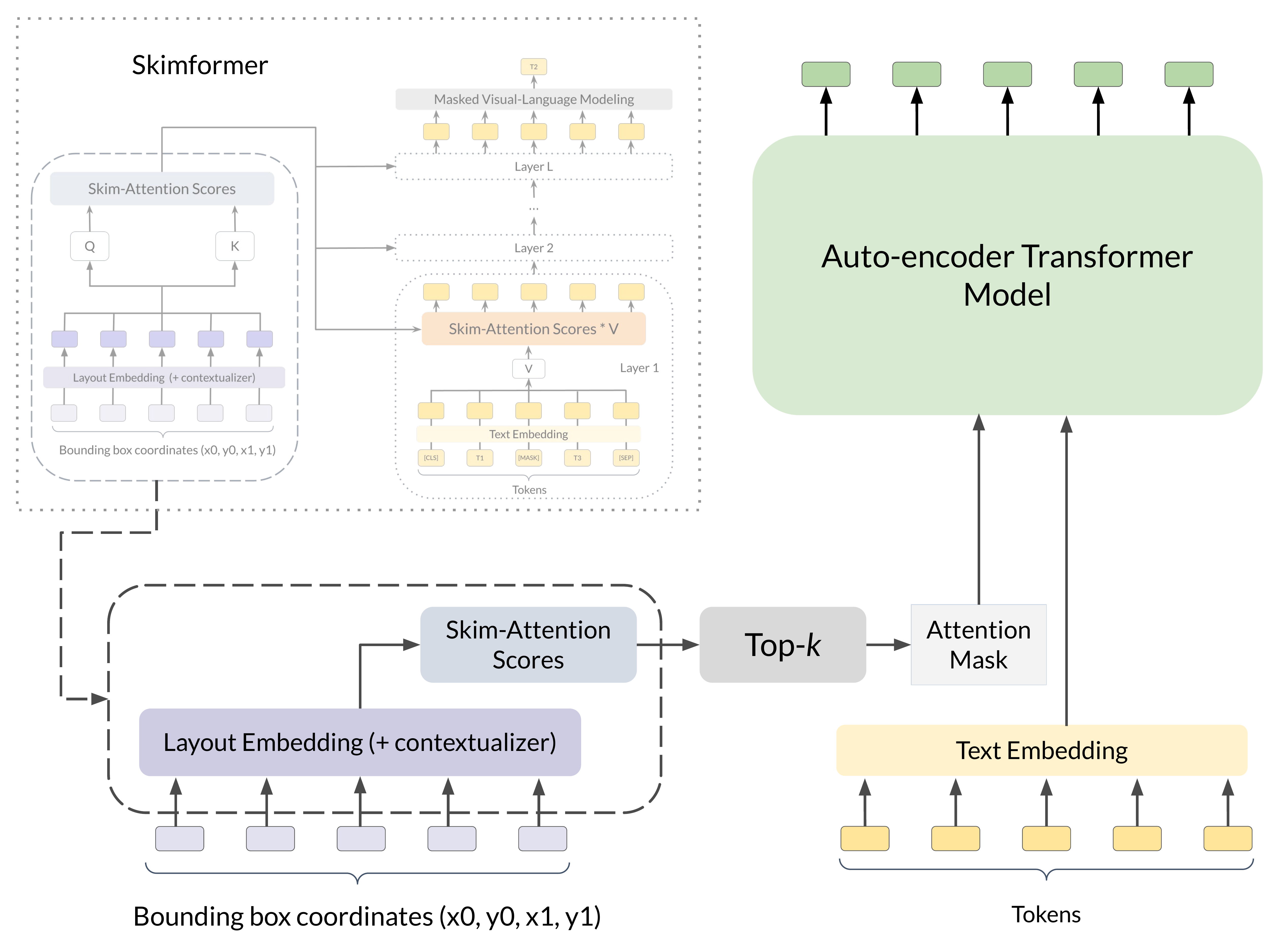}
    \caption{SkimmingMask model architecture. The layout embeddings, Key and Query projections are initialized from an already pre-trained Skimformer model. By filtering the $k$ most attended tokens for each token, the Skim-Attention scores are then converted to an attention mask and given as input to a text-based Transformer model.}
    \label{fig:skimmingmask-architecture}
  \end{subfigure}
  \caption{Our proposed model architectures: Skimformer (left) and SkimmingMask (right). Both models take as input a sequence of tokens and a sequence of token bounding box coordinates.  The input of each modality is converted to an embedding sequence. Only the layout embeddings are used to compute Skim-Attention.}
\end{figure*}

\paragraph{Skimformer} is a two-stage Transformer that replaces self-attention with Skim-Attention. Inspired by previous work in cognitive science, the intuition behind this approach is to mimic how humans process a document by \emph{i)} skimming through the document to extract its structure, and \emph{ii)}  reading the contents informed by the previous step. Skimformer accepts as inputs a sequence of token embeddings and the corresponding sequence of layout embeddings. The model adopts a two-step approach: first, the skim-attention scores are computed once and only once using layout information alone; then, these attentions are used in every layer of a Transformer encoder.
The architecture of Skimformer is depicted in Figure~\ref{fig:skimformer-architecture}.

For a given encoder layer $k$ and a single head, the traditional self-attention operation becomes:

\begin{equation}
\label{eq:skim-attention-full}
	\bm{\alpha}'_k = \textbf{A}^{\ell} \textbf{V}^{t}_{k}
\end{equation}

\noindent where $\textbf{A}^{\ell}$ is the skim-attention matrix obtained through Eq.~\ref{eq:skim-attention-matrix}, and $\textbf{V}^{t}_{k} = \textbf{W}_{v,k} \textbf{X}^t$ is the Value matrix produced by projecting the textual input\footnote{As opposed to BERT, we do not encode sequential positions into the text embeddings.} $\textbf{X}^t = \{\textbf{t}_0, \textbf{t}_1, …, \textbf{t}_n\}$ at layer $k$.

More intuitively, computing skim-attention scores (Eq.~\ref{eq:skim-attention-matrix}) can be interpreted as \textit{skimming through} the document. Information about the semantics (contained in $\textbf{V}$) is then routed based on these similarity scores. This is done via Eq.~\ref{eq:skim-attention-full} and can be seen as \textit{reading} the contents of the document, focusing on the most relevant parts informed by the skim-attention scores.

We train Skimformer using Masked Visual-Language Modeling (MVLM), a pre-training task that extends Masked Language Modeling (MLM) with layout information. MVLM randomly masks some of the input tokens but preserves their layout embeddings. The model is then trained to recover the masked tokens given the contexts. Hence, MVLM helps capture nearby token features, leveraging both semantics and spatial information. 

While we experimented with a standard Transformer model, it is worth noting that any language model can be used as the backbone of Skimformer.

\paragraph{SkimmingMask}

For each token in a sequence, Skim-Attention provides a ranking of the other tokens based on their layout-based similarity. Leveraging this, SkimmingMask uses Skim-Attention as a mask to restrict the computation of self-attention to a smaller number of elements for each token. In this setting, Skim-Attention is viewed as an independent, complementary module that can be plugged into any language model. Given a sequence of layout embeddings, the corresponding skim-attention matrix is converted to an attention mask: based on the similarity scores provided in the attention matrix, each token can only attend to its $k$ most similar tokens. The resulting mask is then given as input to a text-based Transformer language model with standard self-attention, and is used to restrict self-attention for each element in the input text sequence. This can be viewed as sparsifying the standard self-attention matrix.

SkimmingMask is not trainable end-to-end with the Transformer model it is plugged to, as creating an attention mask from an attention matrix is not a differentiable operation (we leave this for future work). Thus, to train this model, the weights for Skim-Attention need to be already trained, and we naturally use the Skimformer weights. The overall architecture of the model is illustrated in Figure~\ref{fig:skimmingmask-architecture}.

We note that SkimmingMask is a new way to cluster tokens: all tokens belonging to the same group have a high similarity to each other regarding their respective \emph{layout position}. This makes SkimmingMask a concurrent approach to Reformer, which reduces the cost of self-attention by clustering tokens into chunks. As opposed to the latter, the concept of similarity is not based on text semantics but on the document structure. Moreover, SkimmingMask does not require the semantic of each token, but only their layout features. Because each token is viewed as a bounding box whose characteristics are only its size and position, the representation space of layout features is much smaller than that of the text, which spans a vocabulary of more than 30k sub-words. As a consequence, computing attention based on layout could require a smaller latent space dimension than for text, corresponding to less computational efforts. This is also the case for humans: as demonstrated in section \ref{section:human-eval}, it is much easier to retrieve information from documents when the layout is provided.

\section{Experiments}

\subsection{Data}

\paragraph{Pre-training Data}

To pre-train our models on a wide variety of document formats, we select three datasets with various non-trivial document layouts: DocBank \citep{li-etal-2020-docbank}, RVL-CDIP \citep{harley2015evaluation} and PubLayNet \citep{zhong2019publaynet}. We combine them by randomly selecting 25k documents from each dataset, for a total of 75K documents. We discard the provided labels and consider these data as unannotated. The resulting dataset is referred to as MIX. As a first evaluation metric, we can compare the perplexity for the different language models on MIX. 

\subparagraph{DocBank}

DocBank is a large-scale dataset that contains 500K English document pages from papers extracted from arXiv.com. These articles span a variety of disciplines (e.g. Physics, Mathematics, and Computer Science), which is beneficial to train more robust models. Pages are split into a training set, validation set and test set with a ratio of 8:1:1. As the authors already extracted the text and bounding boxes using PDFPlumber,\footnote{\url{https://github.com/jsvine/pdfplumber}} there is no need for an OCR system or a PDF parser. To build our subset, we extract 25k document pages: 20k from the full training set, 2,500 from the validation set and 2,500 from the test set. 

\subparagraph{RVL-CDIP}

RVL-CDIP is a large collection of 400k scanned document images from various categories (e.g. letter, form, advertisement, invoice). The wide range of layouts, as well as the low image quality, allows to train more robust models. We select 25k documents from the RVL-CDIP dataset available on Kaggle,\footnote{\url{https://www.kaggle.com/nbhativp/first-half-training}} which amounts to half of the training images from the full dataset (160k images). The text and word bounding boxes are extracted using Tesseract.\footnote{\url{https://github.com/tesseract-ocr/tesseract}} We split the data into 80\% for training, 10\% for validation and 10\% for test.

\subparagraph{PubLayNet}

PubLayNet comprises over 360 thousand document images from PubMed Central\textsuperscript{\texttrademark} Open Access. The medical publications contained in the collection have similar layouts, but the text density coupled with the small image size add to the robustness of the trained models. We extract the first training split among the 7 available on IBM Data Asset eXchange\footnote{\url{https://developer.ibm.com/exchanges/data/all/publaynet/}} and use the first 20k images as our training set. For the validation and test sets, we keep the first 2,500 images in each split. Because OCR accuracy is too low without any pre-processing, we apply a few image processing operations (i.e. rescaling, converting to grayscale, applying dilation and erosion) on each image in order to improve text extraction.

\paragraph{Dataset for Document Layout Analysis}

In addition to perplexity, we evaluate our approach on a downstream task, document layout analysis. Document layout analysis consists in associating each token with its corresponding category: abstract, author, caption, date, equation, footer, list, paragraph, reference, section, table, title and figure.\footnote{We actually discard the \textit{Figure} label, as 1) our models do not take image features into account, and 2) the text associated with such elements is always the same, making the task trivial.}

We use a subset of the full DocBank dataset, created by selecting 10k document pages (distinct from the ones used for pre-training): 8,000 from the full training set, 1,000 from the validation set and 1,000 from the test set. We refer to this dataset as \textit{DocBank-LA}. Each document page is organized as a list of words with bounding boxes, colors, fonts and labels. We use the precision, recall and F1 score defined by \citet{li-etal-2020-docbank}.

\subsection{Experimental Settings}

For reproducibility purposes, we make the code publicly available.\footnote{\url{https://github.com/recitalAI/skim-attention}}

\paragraph{Baselines}

We compare our models with three baselines: i) the text-only BERT, ii) the multi-modal LayoutLM, and iii) the text-only Longformer for long documents. Note that the LayoutLM architecture is based on BERT, with additional layout components. For fair comparison, all our models designed for short sequences are based on BERT as well, as detailed below. 

\paragraph{Pre-training}

For BERT, LayoutLM and Longformer, we use their default architecture. Following the BERT base model, Skimformer consists of a 12-layer Transformer encoder with 12 attention heads and a hidden size set to 768 for both text and layout embeddings, amounting to 99M parameters. We further add a 2-layer Transformer encoder to contextualize the layout embeddings, which increases the number of parameters to 113M. To test Skim-Attention on longer documents, we build LongSkimformer, a combination of Skim-Attention and Longformer. Every model is trained from scratch on the MIX dataset for 10k steps. We set the maximum sequence length to $n = 512$ for every model except for Longformer and LongSkimformer, for which $n = 2,048$. Skimformer, LongSkimformer and LayoutLM are pre-trained using MVLM, while BERT and Longformer are pre-trained with MLM. For more implementation details, see Section ~\ref{supp-sec:implementation-details} in the appendix.

\paragraph{Document Layout Analysis}

As DocBank contains fine-grained token-level annotations, we consider the document layout analysis task as a sequence labeling task. Each model pre-trained on MIX is fine-tuned on this downstream task for 10 epochs. Section \ref{supp-sec:implementation-details} of the appendix provides a detailed description of the settings used. For the SkimmingMask models, we selected the hyperparameter $k$ on validation, i.e. the number of tokens that can be attended to.\footnote{We tested $k \in [512, 384, 256, 128]$.}

\subsection{Results and Discussion}

\subsubsection{Perplexity}

\begin{table}
\centering \small
\begin{tabular}{lr}
    \hline
    \textbf{Model} & \textbf{Test Perplexity}\\
    \hline
    BERT \citep{devlin-etal-2019-bert} &  357.11 \\
    LayoutLM \citep{xu2020layoutlm}    & 45.86 \\
    Skimformer                         & 33.77 \\
    \midrule
    Longformer \citep{beltagy2020longformer} & 333.28 \\
    LongSkimformer                     & \textbf{32.02} \\
    \hline
\end{tabular}
\caption{Test perplexity on the MIX dataset after 10k optimization steps. Each model was trained from scratch. Bold denotes the best score.}
\label{tab:ppl-mix}
\end{table}

\begin{figure}
    \centering \small
    \includegraphics[width=.4\textwidth]{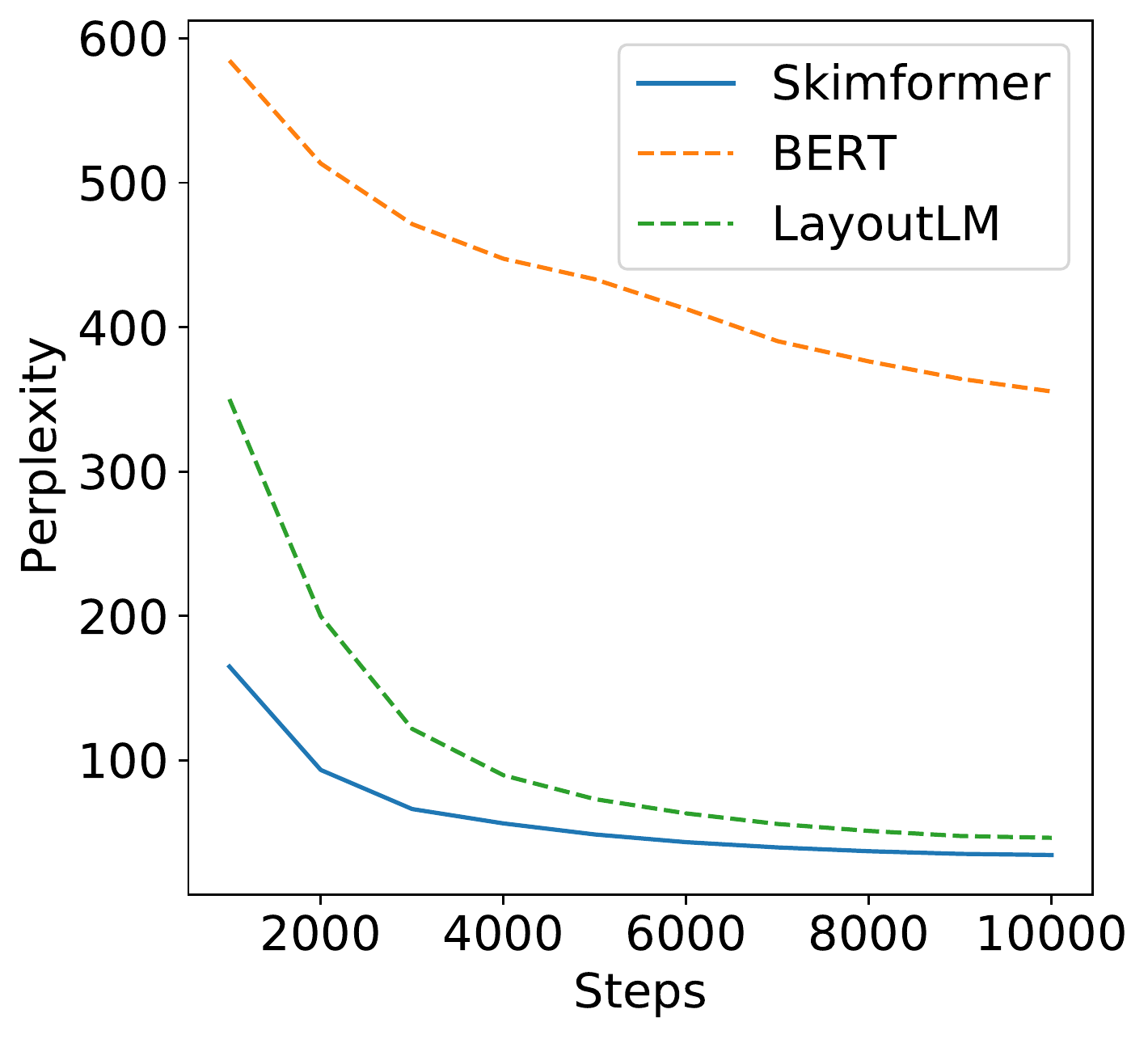}
    \caption{Model perplexity on the MIX validation set with respect to the number of optimization steps. All models are trained from scratch.}
    \label{fig:pretraining-learning-curves}
\end{figure}

In Table~\ref{tab:ppl-mix}, we report the perplexity on the MIX dataset. We observe that Skimformer and LongSkimformer respectively outperform BERT and Longformer by a huge margin, while improving perplexity by more than 10 points over LayoutLM. In addition, Figure~\ref{fig:pretraining-learning-curves} demonstrates that Skimformer converges much faster than BERT, and slightly more than LayoutLM.

\begin{table}[t]
\centering \small
\begin{tabular}{lr}
    \hline
    \textbf{Skim-Attention Input} & \textbf{Test Perplexity}\\
    \hline
    Layout                              & 36.41 \\
    1D position                         & 54.39 \\ 
    Uniform layout                      & 421.97 \\
    Degraded layout                     & 103.39 \\
    Contextualized layout               & \textbf{33.77} \\
    \hline
\end{tabular}
\caption{Ablation study on the MIX dataset, where perplexity on the test set is reported. All models were trained from scratch. Bold denotes the best score.}
\label{tab:ablation-study}
\end{table}

\paragraph{Ablation Study}

We further conduct an ablation study about the influence of the Skim-Attention inputs on Skimformer's performance. The results are listed in Table~\ref{tab:ablation-study}. To estimate the impact of the input type, we consider a Skimformer model i) wherein Skim-Attention is based on sequential positions (1D position), ii) the bounding boxes are all set to the same fixed value, preventing the model to gather any information about the true location (Uniform layout), iii) they are replaced by their centers (Degraded layout), and iv) the layout embeddings are contextualized (Contextualized Layout). 

We can see that replacing spatial with sequential positions results in an increase in perplexity, indicating that layout information is crucial for the Language Model. It is also observed that assigning the same bounding box to every token leads to a severe drop in performance. Coupled with the perplexity obtained with a degraded layout, this shows that the model's performance is greatly impacted by the layout input quality. At last, contextualizing the layout inputs through a small Transformer brings slight improvements over computing Skim-Attention directly on the layout embeddings.

Finally, we benchmark Skimformer and LayoutLM on both speed and peak memory usage for training. Results provided in Figure~\ref{supp-fig:benchmark} of the appendix show that Skimformer is more time and memory efficient than LayoutLM.

\subsubsection{Document Layout Analysis}

\begin{table*}
\centering
\small
\begin{threeparttable}
\begin{tabular}{lcccccccc}
    \toprule
     & \textbf{Skimming} & \textbf{Seq.} & \multicolumn{2}{c}{\textbf{Nb Attentions}} & \textbf{Total} & & & \\
    \textbf{Model} & \textbf{Mask} & \textbf{Len} & Original\tnote{*} & Skim-Attn & \textbf{Compute} & \textbf{Rec.} & \textbf{Prec.} & \textbf{F1} \\
    \midrule
    BERT \citep{devlin-etal-2019-bert}  & \xmark                         & 512 & 12              & 0 & 100.00\% & 67.21 & 59.28 & 60.98 \\ 
    LayoutLM \citep{xu2020layoutlm}        & \xmark                         & 512 & 12              & 0 & 100.00\% & 81.60 & 77.96 & \textbf{79.28} \\
    \midrule 
    Skimformer          & \xmark                         & 512 & 0 & 3\tnote{**} & 25.00\% & 78.80 & 74.35 & 75.86 \\
    BERT+SkimEmbeddings & \xmark                         & 512 & 12              & 0 & 100.00\% & \textbf{82.42} & 77.06 & \textbf{79.16} \\
    
    BERT+$\textrm{SkimmingMask}$                & \cmark  & 128  & 12  & 3\tnote{**} & 31.25\% & 72.32 & 64.39 & 67.36 \\
    LayoutLM+$\textrm{SkimmingMask}$              & \cmark & 128 & 12 & 3\tnote{**} & 31.25\% & 81.15 & \textbf{78.30} & \textbf{79.26} \\ 
    \midrule 
    Longformer \citep{beltagy2020longformer} & \xmark & 2,048 & 12 & 0 & 100\% & 74.88 & 69.29 & 71.17 \\
    LongSkimformer & \xmark & 2,048 & 0 & 3\tnote{**} & 25\% & 81.22 & 73.45 & 76.61 \\
\bottomrule
\end{tabular}
\begin{tablenotes}
  \item[*] Standard self-attention for Skimformer, BERT-based and LayoutLM-based models. Longformer self-attention for Longformer and LongSkimformer.
  \item[**] Attention is computed twice (by a 2-layer Transformer) during layout contextualization, then once by Skim-Attention.
\end{tablenotes}
\end{threeparttable}
\caption{Model performance (in \%) on the DocBank-LA dataset. \textit{Seq. Len} indicates the number of tokens attended with either standard attention (for Skimformer, BERT-based and LayoutLM-based models), or Longformer attention (for Longformer and LongSkimformer). \textit{Nb Attention} represents the number of times attention (original and Skim-Attention) is computed and stored. \textit{Total Compute} specifies the ratio of the final computational cost (\# operations needed to compute attention) w.r.t. BERT/LayoutLM or Longformer. Each model was pre-trained from scratch on MIX, then fine-tuned on DocBank-LA. }
\label{fig:results-docbank}
\end{table*}

Table~\ref{fig:results-docbank} reports the performance on DocBank-LA, the sequence length processed, the number of times attention is computed and the ratio of the total calculation unit ($n^2 \times \textrm{Nb Skim-Attn} + \textrm{Seq. Len}^2 \times \textrm{Nb Standard Attn}$, where $n$ is the length of the initial sequence on which Skim-Attention is applied; and \textit{Seq. Len} is the length obtained after applying SkimmingMask) to that of BERT/LayoutLM and Longformer. All models were pre-trained from scratch on MIX. 

Skimformer is substantially superior to BERT, improving the F1 score by 15\% while reducing the number of attentions computed by four. We experimented with plugging the layout embeddings learnt by Skimformer in a BERT model. The resulting model, BERT+SkimEmbeddings, resembles LayoutLM in terms of architecture.\footnote{In BERT+SkimEmbeddings, the layout embeddings are first projected into the same dimensional space as the text embeddings. In this way, we can plug the layout embeddings from any Skimformer model, in particular smaller ones.} Results show that BERT+SkimEmbeddings performs on par with LayoutLM despite simply combining separately pre-trained modalities, as opposed to the latter which requires an extensive joint training.

For the SkimmingMask models (see the last two rows in Table~\ref{fig:results-docbank}), the models attend to only the top-$k$ 128 tokens. Compared to LayoutLM, this reduction to the quadratic factor allows to obtain the same downstream results with only 31.25\% of the computational burden.
Compared to BERT, it even obtains an absolute improvement of more than 6\% in term of F1 score.

LongSkimformer benefits from both Skim-Attention and Longformer's gain in efficiency. It outperforms Longformer by 5\% while requiring four times less attention operations, and the use of Longformer's linear attention allows LongSkimformer to process sequences four times larger than Skimformer can.

\subsection{Attention Visualization}

\begin{figure*}
    \centering \small
    \includegraphics[width=0.6\textwidth]{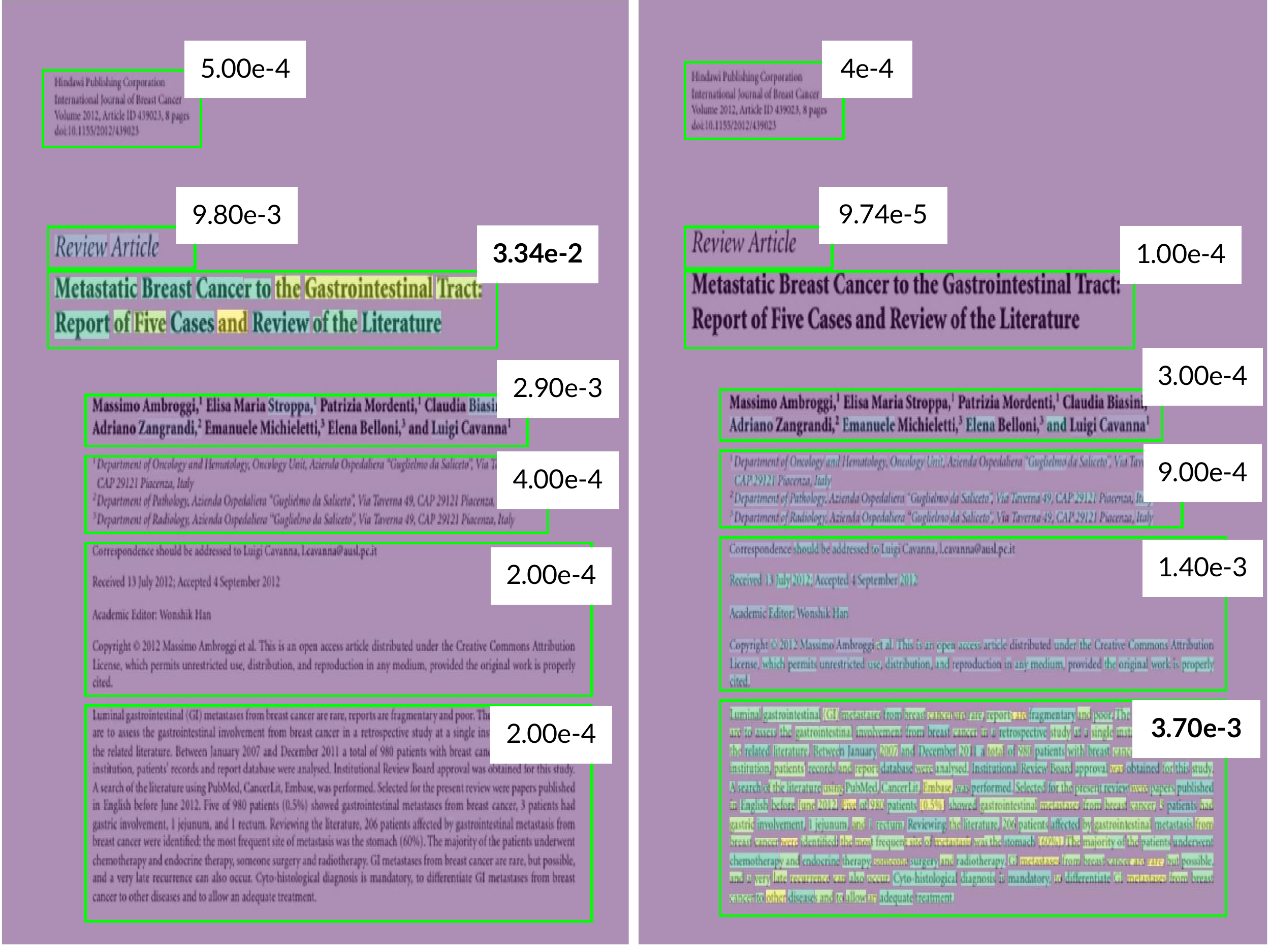}
    \caption{Skim-attention maps corresponding to the title (left) and the abstract (right), along with average attention score (in white) per text block (in green). We consider the skim-attention matrix averaged over all the attention heads. Given a semantic unit (title or abstract), we plot the average attention score for each token.
    }
    \label{fig:attention-vis}
\end{figure*}

Figure~\ref{fig:attention-vis} shows the attention maps produced by Skimformer on a sample document.\footnote{Randomly picked. Similar results can be observed across different samples (see Section~\ref{supp-fig:attention-vis} in the appendix).} Given a semantic unit (either title or abstract in our example), we select the corresponding tokens and compute their average attention over the whole document. We observe, both qualitatively and quantitatively, that tokens attend mainly to other elements in the same semantic unit, thus creating clusters of tokens that are relevant to each other. This shows that the model has grasped the concept of semantic unit with only self-supervision, enabling the emergence of a document structure representation. We argue that these structure-aware clusters could pave the way for long text encoding and unsupervised document segmentation. 

\section{Conclusion}
We present Skim-Attention, a new structure-aware attention mechanism.  We conduct extensive experiments to show the effectiveness of Skim-Attention, both as an end-to-end model (Skimformer) and as a mask for any language model (Skim-Attention). We hope this work will pave the way towards a new research direction for efficient attentions. For future works, we will investigate how to integrate image features, and explore tasks that require capturing longer-range dependencies.

\bibliographystyle{acl_natbib}
\bibliography{refs}

\clearpage 
\appendix

\begin{center}{
\Large
\textbf{Skim-Attention: Learning to Focus via Document Layout -- Appendix}
}
\end{center}

\section{Preliminary Experiments: Human Evaluation}
\label{supp-section:human-eval}

\begin{figure*}[!htbp]
\centering
  \begin{subfigure}[b]{0.49\textwidth}
    \includegraphics[width=\textwidth]{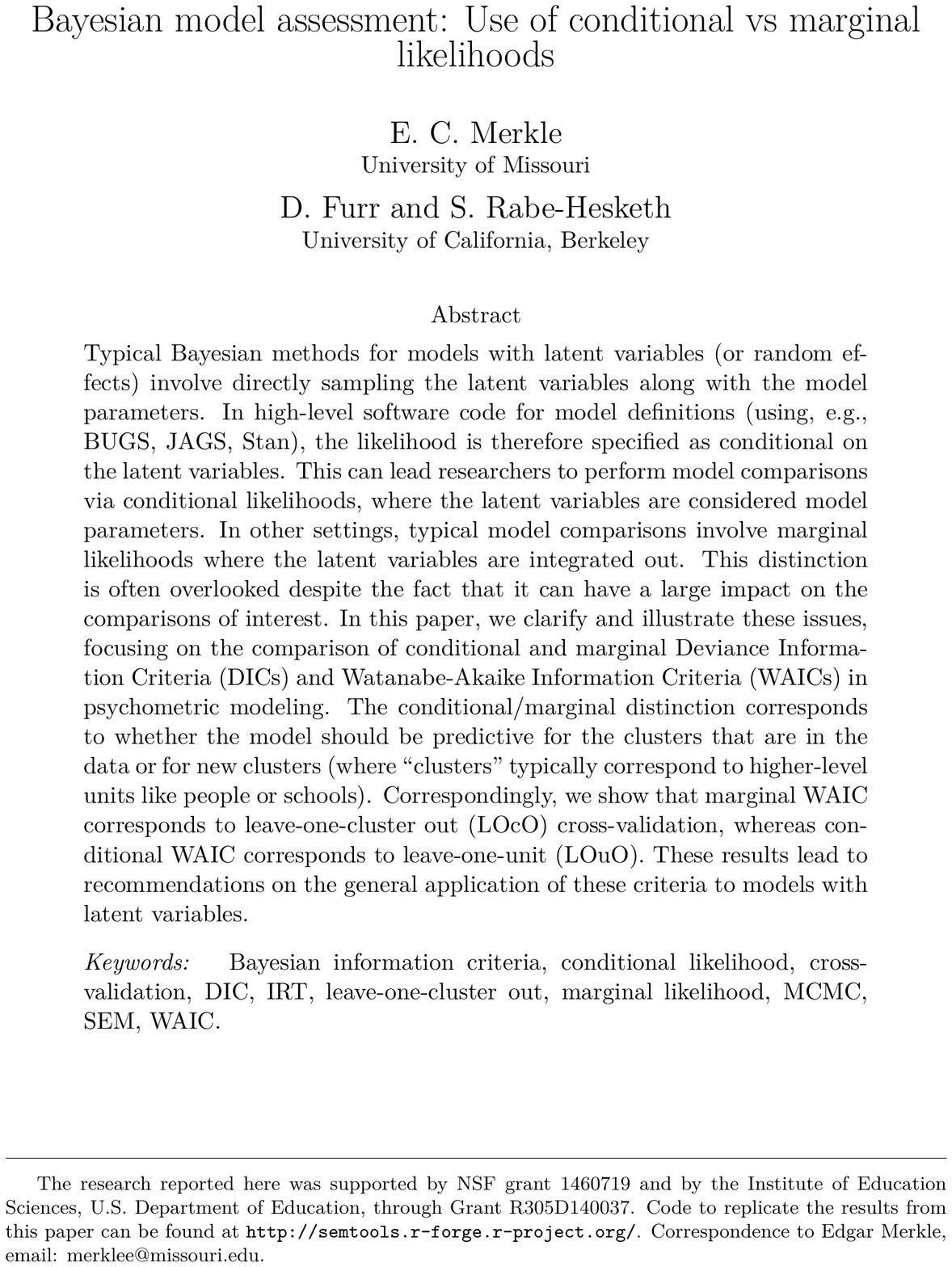}
    \caption{}
  \end{subfigure}
  \hfill
  \begin{subfigure}[b]{0.49\textwidth}
    \includegraphics[width=\textwidth]{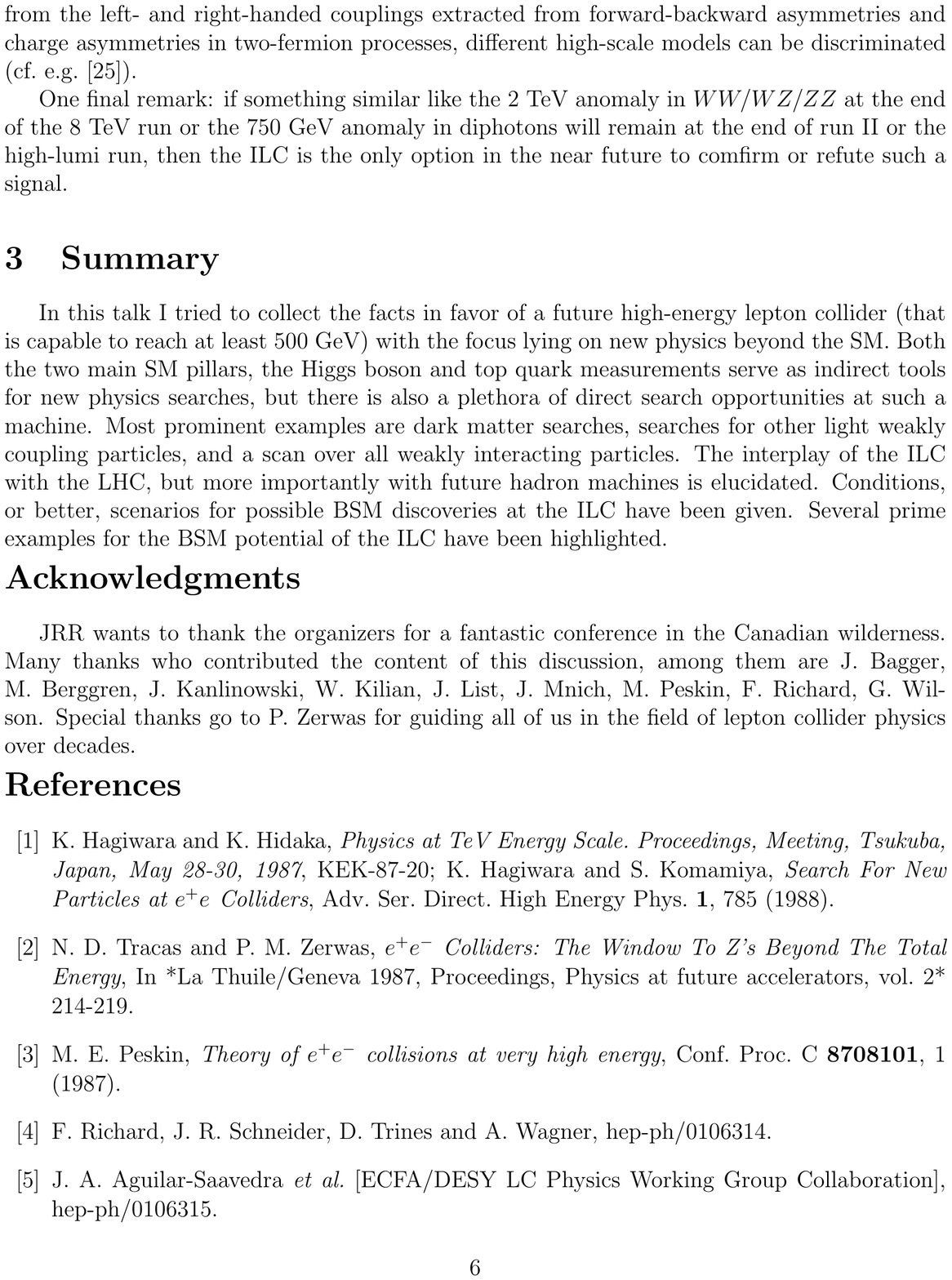}
    \caption{}
  \end{subfigure}
  \begin{subfigure}[b]{0.49\textwidth}
    \includegraphics[width=\textwidth]{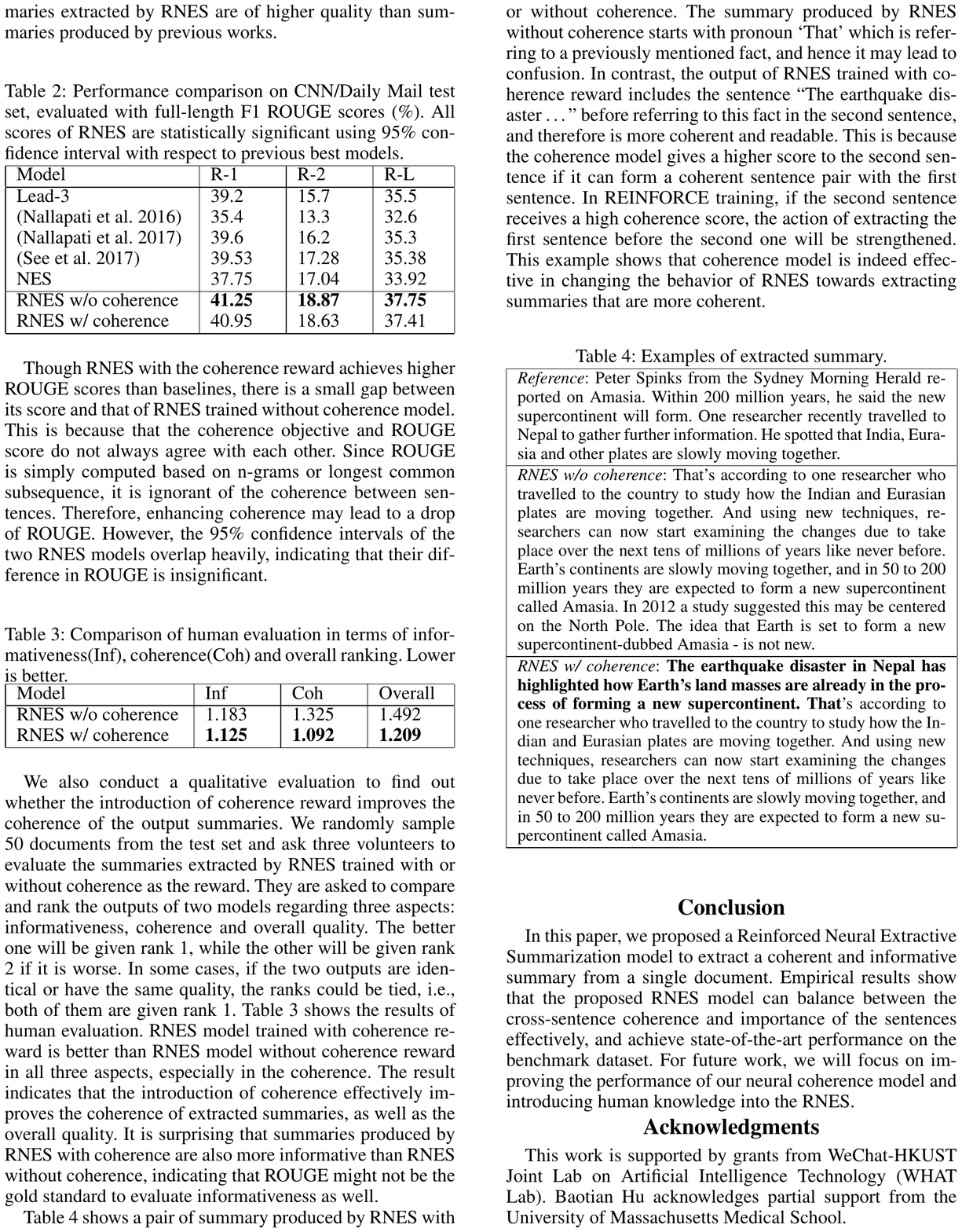}
    \caption{}
  \end{subfigure}
  \begin{subfigure}[b]{0.49\textwidth}
    \includegraphics[width=\textwidth]{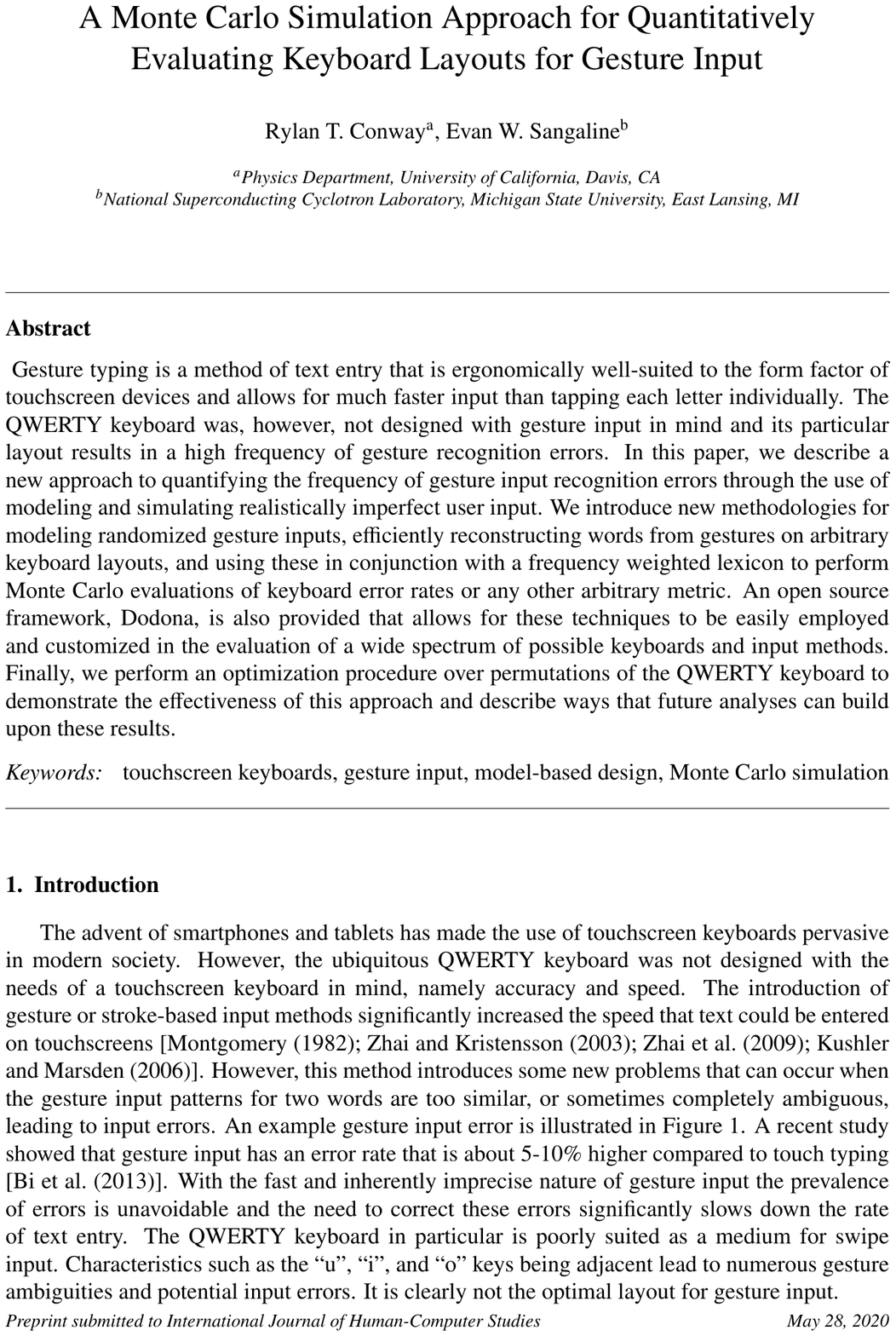}
    \caption{}
  \end{subfigure}
  \caption{Documents selected for our preliminary cognitive experiment.}
  \label{supp-fig:samples-human-eval}
\end{figure*}

To evaluate the impact of layout on readers' understanding, we conduct an experiment in which we measure the amount of time required for human annotators to answer questions from both formatted and non-formatted documents. We hand-pick four document pages from the DocBank dataset \citep{li-etal-2020-docbank}, and create a plain-text version out of each of these documents by flattening them. This results in eight pages: the four original document pages, and their serialized versions with no layout nor formatting. The original, formatted document pages are displayed in figure \ref{supp-fig:samples-human-eval}. We create two basic questions for each document (answers are provided in italic):

\begin{itemize}
    \item Document (a) :
    \begin{itemize}
        \item Who are the authors of this paper ? \textit{E.C Merkle, D. Furr, S. Rabe-Hesketh.}
        \item What are the keywords ? \textit{Bayesian information criteria, conditional likelihood, cross-validation, DIC, IRT, leave-one-cluster out, marginal likelihood, MCMC, SEM, WAIC.}
    \end{itemize}
    \item Document (b) :
    \begin{itemize}
        \item What paper did N.D. Tracas and P.M. Zerwas write ? \textit{e + e – Colliders: The Window To Z’s Beyond The Total Energy.}
        \item Who does the author thank ? \textit{The organizers, those who contributed to the content of the discussion (J. Bagger, M. Berggren, J. Kanlinowski, W. Kilian, J. List, J. Mnich, M. Peskin, F. Richard, G. Wilson), P. Zerwas.}
    \end{itemize}
    \item Document (c) :
    \begin{itemize}
        \item What is proposed in this paper ? \textit{A Reinforced Neural Extractive Summarization model to extract a coherent and informative summary from a single document.}
        \item What is compared in table 3 ? \textit{Human evaluation in terms of informativeness(Inf), coherence(Coh) and overall ranking.}
    \end{itemize}
    \item Document (d) :
    \begin{itemize}
        \item When was this paper submitted ? \textit{May 2028, 2020.}
        \item What are the keywords of this paper ? \textit{Touchscreen keyboards, gesture input, model-based design, Monte Carlo simulation.}
    \end{itemize}
\end{itemize}

Four annotators are asked to answer these questions. Each of them alternates between fully formatted contents (i.e. the original document page) and plain text. We decide that annotators 1 and 3 have access to the document layout for documents 1 and 3, while annotators 2 and 4, for documents 2 and 4.

Given a document, the instructions are as follows:

\begin{enumerate}
    \item Read the entire document, then the questions;
    \item Start the timer;
    \item Find the answer to the first question (without writing it down);
    \item Stop the timer and check if the answer is correct;
    \begin{itemize}
        \item If this is the case, write down the time indicated by the timer, then reset it and answer the second question by re-iterating steps 2 to 4.
        \item If not, resume timer until you find the correct answer.
    \end{itemize}
    \item Proceed to the next document.
\end{enumerate}

\begin{table}[!htbp]
\centering
\small
\begin{tabular}{lcc}
    \hline
                & \textbf{Formatted} & \textbf{Plain-text} \\
    \hline
    \textbf{Doc 1} & 2.95 $\pm$ 0.43 & 10.85 $\pm$ 5.67 \\
    \textbf{Doc 2} & 4.29 $\pm$ 1.54 & 14.44 $\pm$ 11.25 \\
    \textbf{Doc 3} & 7.75 $\pm$ 0.38 & 20.81 $\pm$ 10.26 \\
    \textbf{Doc 4} & 9.20 $\pm$ 4.57 & 14.65 $\pm$ 9.06 \\ 
    \hline
\end{tabular}
\caption{Time (in seconds) required to retrieve information per document and document type (formatted/non-formatted). Standard deviation is also reported.}
  \label{supp-tab:human-eval-avg-results}
\end{table}

\begin{table*}[!htbp]
\centering
\small
\begin{tabular}{ccccccccccccc}
    \hline
    & \multicolumn{3}{c}{\textbf{Doc 1}} & \multicolumn{3}{c}{\textbf{Doc 2}} & \multicolumn{3}{c}{\textbf{Doc 3}} & \multicolumn{3}{c}{\textbf{Doc 4}} \\ 
    \hline 
    & \textbf{Q1} & \textbf{Q2} & \textbf{AVG} & \textbf{Q1} & \textbf{Q2} & \textbf{AVG} & \textbf{Q1} & \textbf{Q2} & \textbf{AVG} & \textbf{Q1} & \textbf{Q2} & \textbf{AVG} \\
    \hline
    \textbf{A 1} & \cellcolor{yellow!25} 1.48 & \cellcolor{yellow!25} 3.81 & \cellcolor{yellow!25} 2.65 & 19.52 & 25.27 & 22.40 & \cellcolor{yellow!25} 11.91 & \cellcolor{yellow!25} 4.12 & \cellcolor{yellow!25} 8.02 & 12.98 & 29.12 & 21.05 \\
    \textbf{A 2} & 9.23 & 4.44 & 6.84 & \cellcolor{yellow!25} 5.12 & \cellcolor{yellow!25} 5.63 & \cellcolor{yellow!25} 5.38 & 24.5 & 31.62 & 28.06 & \cellcolor{yellow!25} 7.02 & \cellcolor{yellow!25} 4.92 & \cellcolor{yellow!25} 5.97 \\
    \textbf{A 3} & \cellcolor{yellow!25} 3.76 & \cellcolor{yellow!25} 2.76 & \cellcolor{yellow!25} 3.26 & 7.19 & 5.77 & 6.48 & \cellcolor{yellow!25} 9.19 & \cellcolor{yellow!25} 5.77 & \cellcolor{yellow!25} 7.48 & 3.93 & 12.55 & 8.24 \\
    \textbf{A 4} & 8.66 & 21.05 & 14.86 & \cellcolor{yellow!25} 3.49 & \cellcolor{yellow!25} 2.9 & \cellcolor{yellow!25} 3.20 & 15.9 & 11.2 & 13.55 & \cellcolor{yellow!25} 6.9 & \cellcolor{yellow!25} 17.96 & \cellcolor{yellow!25} 12.43\\
    \hline
\end{tabular}
\caption{Time (in seconds) taken by each annotator to answer each question. Average per document is also reported. Yellow cells indicate that the document layout was provided for corresponding documents and annotators.}
 \label{supp-tab:human-eval-full-results}
\end{table*}

Results per document and document type (i.e., formatted or non-formatted) are given in table \ref{supp-tab:human-eval-avg-results}. The entirety of the results is reported in table \ref{supp-tab:human-eval-full-results}.

\section{Implementation Details}
\label{supp-sec:implementation-details}

\paragraph{Pre-training}

For BERT, LayoutLM and Longformer, we use the PyTorch implementation from Hugging Face's Transformers library \citep{wolf-etal-2020-transformers}.

Each model is trained from scratch on the MIX dataset for 10k steps with a batch size of 8, except for Longformer which was trained with a smaller batch size of 4 due to memory limitations. We use the Adam optimizer with weight decay fix \citep{loshchilov2017decoupled}, a weight decay of 0.01 and $(\beta_1, \beta_2) = (0.9, 0.999)$. The learning rate is set to $1e^{-4}$ and linearly warmed up over the first 100 steps. The maximum sequence length is set to $n = 512$, with the exception of Longformer and LongSkimformer, for which $n=2,048$. Following BERT, we mask 15\% of the text tokens in MVLM, among which 80\% are replaced by a special token [MASK], 10\% are replaced by a random token, and 10\% remains the same.

\paragraph{Document Layout Analysis}

Each model pre-trained on MIX is fine-tuned on DocBank's document layout analysis task for 10 epochs, with a learning rate of $5e^{-5}$ and a batch size of 8 (except for Longformer which was fine-tuned with a batch size of 4). The models are extended with a token-classification head on top, consisting of a linear layer followed by a softmax layer, and are trained using cross-entropy. 

For SkimmingMask, we select the Skim-Attention module from the Skimformer model pre-trained from scratch on MIX. We then plug it into a BERT model, also pre-trained from scratch on MIX. The resulting model is fine-tuned with the same settings as described previously.

\section{Benchmark}

Using Hugging Face's Transformers benchmarking tools \citep{wolf-etal-2020-transformers}, we benchmark Skimformer and LayoutLM on both speed and required memory for pre-training. We consider the base variant of LayoutLM, and use the implementation from the Transformers library. In addition to the full Skimformer, we evaluate a variant in which the small Transformer contextualizing layout embeddings is removed (Skimformer-no-context). The batch size is fixed to 8, and memory and time performance is evaluated for the following sequence lengths: 8, 32, 128 and 512. We use Python 3.7.10, PyTorch 1.8.1+cu101 \citep{paszke2019pytorch}, and Transformers 4.6.0.dev0. All experiments were conducted on one Tesla T4 with 15GB of RAM.

\begin{figure*}[!htbp]
\centering
\small
  \begin{subfigure}[b]{0.49\textwidth}
    \includegraphics[width=\textwidth]{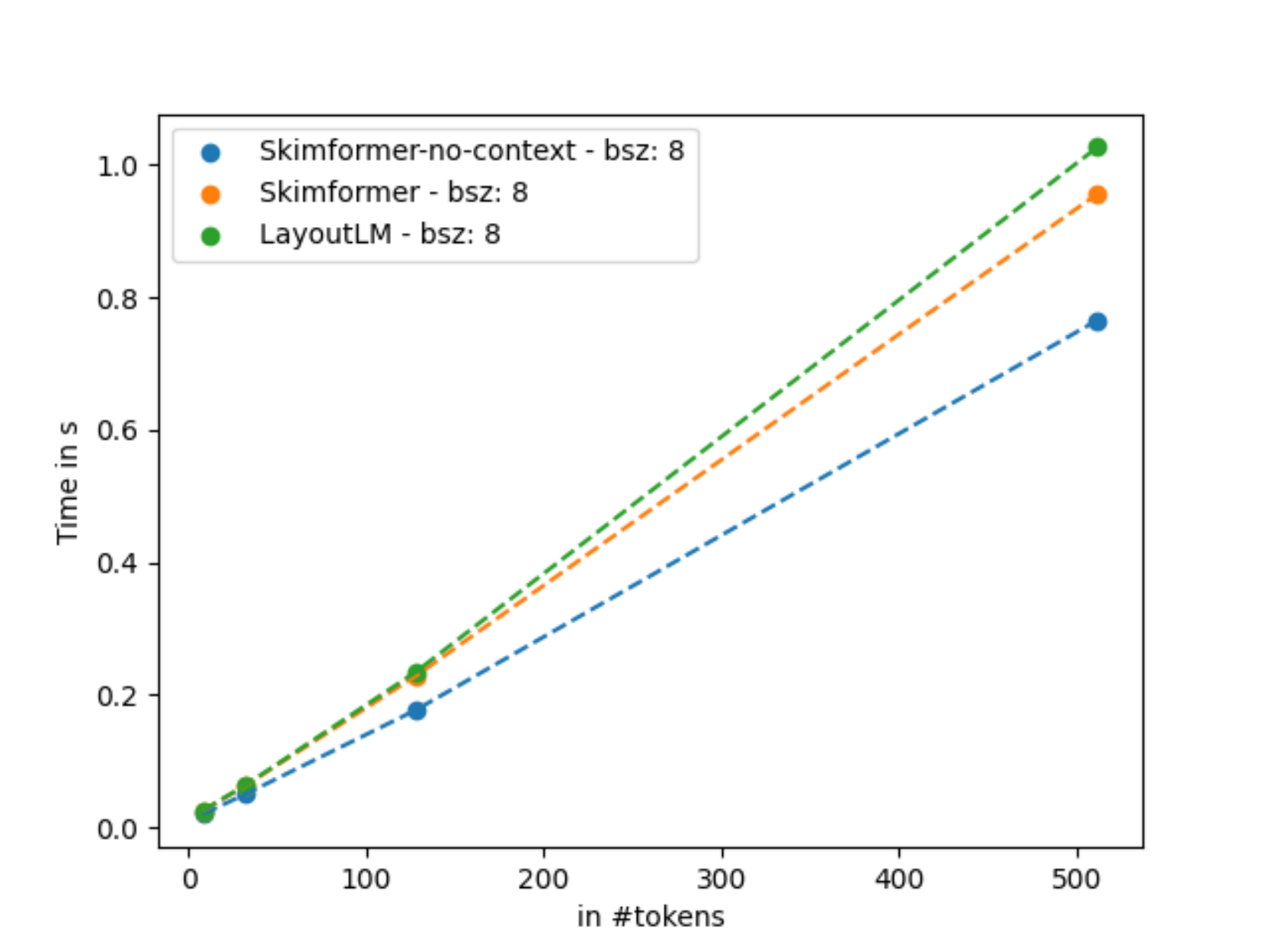}
    \caption{Time usage for pre-training.}
    \label{supp-fig:benchmark-train-time}
  \end{subfigure}
  \begin{subfigure}[b]{0.49\textwidth}
    \includegraphics[width=\textwidth]{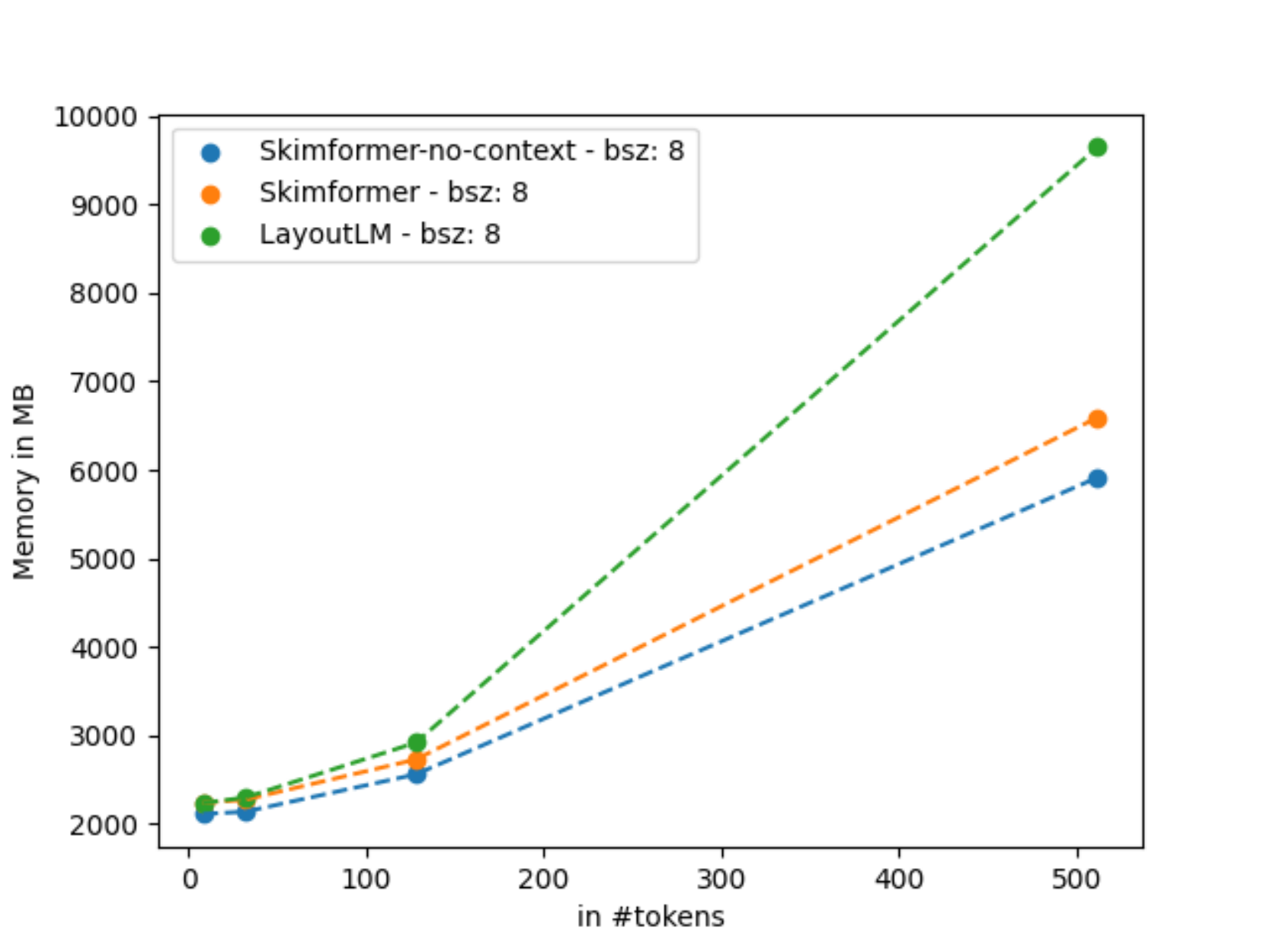}
    \caption{Memory usage for pre-training.}
    \label{supp-fig:benchmark-train-memory}
  \end{subfigure}
  \caption{Comparison of time and memory usage for LayoutLM (green), Skimformer with layout contextualizer (orange) and without (blue). Results are plotted against sequence length.}
  \label{supp-fig:benchmark}
\end{figure*}

Figure \ref{supp-fig:benchmark-train-memory} reports the time (figure \ref{supp-fig:benchmark-train-time}) and peak memory consumption (figure \ref{supp-fig:benchmark-train-memory}) with respect to the sequence length.

\section{Attention Visualization}

\begin{figure*}[!htbp]
\centering
  \begin{subfigure}[b]{\textwidth}
    \hspace{2.2cm}
    \includegraphics[width=1.5\textwidth]{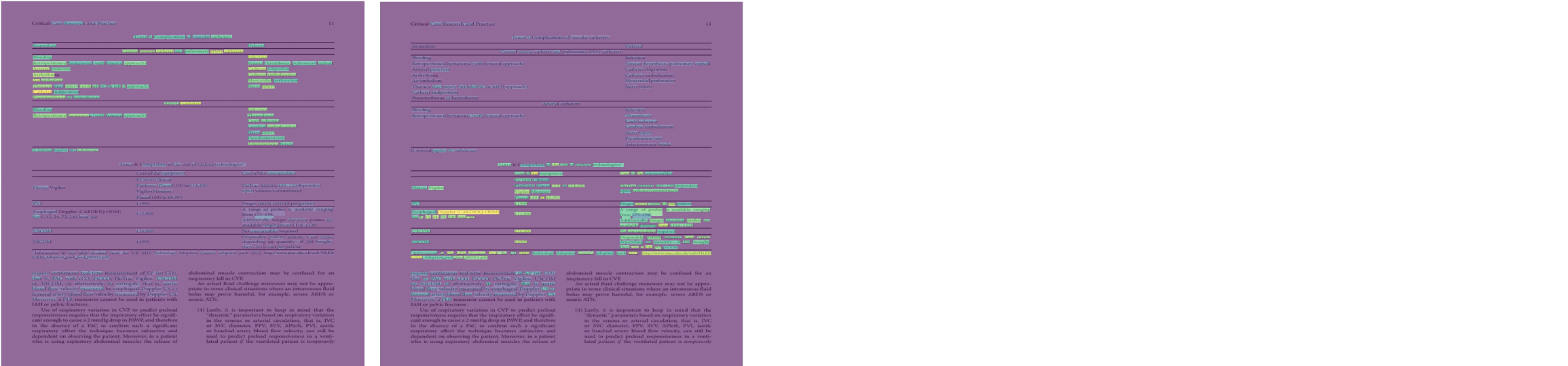}
    \caption{Skim-attention maps corresponding to the top table (left) and the bottom table (right).}
    \label{supp-fig:attention-vis-tables}
  \end{subfigure}
  
  \begin{subfigure}[b]{\textwidth}
    \hspace{-1.5cm}
    \includegraphics[width=1.2\textwidth]{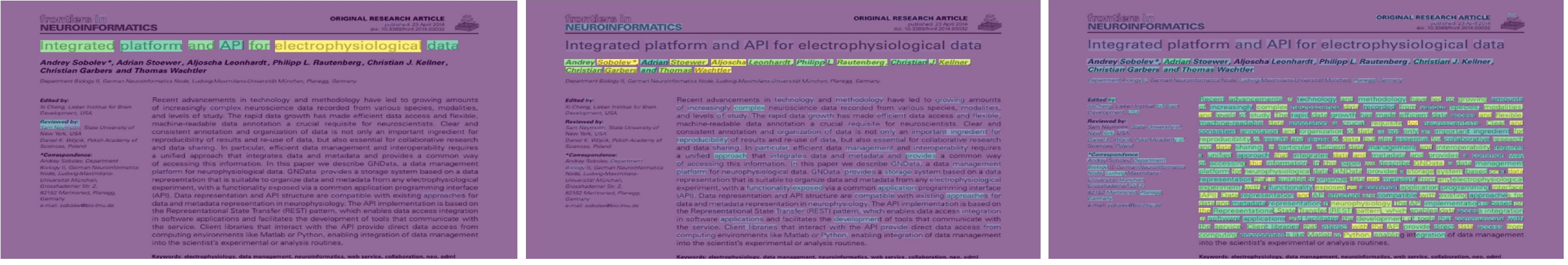}
    \caption{Skim-attention maps corresponding to the title (left), the authors (center) and the abstract (right).}
    \label{supp-fig:attention-vis-firstpage}
  \end{subfigure}
  \caption{Skim-Attention maps on two sample documents.}
  \label{supp-fig:attention-vis}
\end{figure*}

Figure \ref{supp-fig:attention-vis} contains the attention maps obtained by Skimformer on two documents sampled from PubLayNet \citep{zhong2019publaynet}. For each sample, we average the attention scores of tokens belonging to a given semantic unit, and map the result to the document image. In the first document (figure \ref{supp-fig:attention-vis-tables}), we focus on the top table (left) and the bottom one (right). In the second sample (figure \ref{supp-fig:attention-vis-firstpage}), we investigate the title (left), the authors (center) and the abstract (right).

\end{document}